\documentclass[letterpaper]{article} 
\usepackage{aaai24}  
\usepackage{times}  
\usepackage{helvet}  
\usepackage{courier}  
\usepackage[hyphens]{url}  
\usepackage{graphicx} 
\usepackage{xcolor}
\usepackage{natbib}  
\usepackage{caption} 
\usepackage{algorithm}
\usepackage{graphicx}
\usepackage{subfigure}
\usepackage{adjustbox}
\usepackage{amsmath}
\usepackage{amssymb}
\usepackage{booktabs}
\usepackage{algpseudocode}

\begin{document}

\title{Traffic Signal Control Using Lightweight Transformers: \\An Offline-to-Online RL Approach}
\author{
    Xingshuai Huang,
        Di Wu, 
        and~Benoit Boulet 
}
\affiliations{
    Department of Electrical and Computer Engineering \\ McGill University
}

\maketitle

\begin{abstract}

Efficient traffic signal control is critical for reducing traffic congestion and improving overall transportation efficiency. The dynamic nature of traffic flow has prompted researchers to explore Reinforcement Learning (RL) for traffic signal control (TSC). Compared with traditional methods, RL-based solutions have shown preferable performance. However, the application of RL-based traffic signal controllers in the real world is limited by the low sample efficiency and high computational requirements of these solutions. In this work, we propose DTLight, a simple yet powerful lightweight Decision Transformer-based TSC method that can learn policy from easily accessible offline datasets. DTLight novelly leverages knowledge distillation to learn a lightweight controller from a well-trained larger teacher model to reduce implementation computation. Additionally, it integrates adapter modules to mitigate the expenses associated with fine-tuning, which makes DTLight practical for online adaptation with minimal computation and only a few fine-tuning steps during real deployment.
Moreover, DTLight is further enhanced to be more applicable to real-world TSC problems. Extensive experiments on synthetic and real-world scenarios show that DTLight pre-trained purely on offline datasets can outperform state-of-the-art online RL-based methods in most scenarios. Experiment results also show that online fine-tuning further improves the performance of DTLight by up to 42.6\% over the best online RL baseline methods. In this work, we also introduce \textit{D}atasets specifically designed for \textit{T}SC with offline \textit{RL} (referred to as DTRL). Our datasets and code are publicly available~\footnote{https://github.com/XingshuaiHuang/DTLight}.

\end{abstract}

\section{Introduction}

Traffic congestion has led to significant negative impacts on Urban areas, resulting in adverse economic and environmental consequences, such as increased travel time, fuel consumption, and air pollution~\citep{yang2023hierarchical}. This issue is expected to get worse as the global population continues to grow, leading to greater demand for transportation. Traffic signal control is a promising solution that requires minimal infrastructure reconstruction to manage traffic flow and reduce congestion. However, conventional TSC techniques, such as fixed-time \citep{roess2004traffic, ault2021reinforcement} and rule-based \citep{varaiya2013max, xu2022integrating} 
controllers, rely on preset timings or pre-defined rules, making them incapable of handling dynamic traffic circumstances. Reinforcement learning \citep{sutton2018reinforcement} allows controllers to adapt the decisions based on the environment states, making it a potent data-driven solution.

Researchers have proposed numerous RL-based TSC approaches~\citep{ault2021reinforcement, zhu2023metavim, yang2023hierarchical, qiao2023traffic, zhu2023multi}. 
Studies include the design of Markov decision processes (MDP) for TSC~\citep{wu2021efficient, zhang2022expression}, collaborative control of traffic signals~\citep{jiang2022multi, yang2023hierarchical, qiao2023traffic}, and more~\citep{jiang2021dynamic, du2023safelight}.
In general, reinforcement learning-based methods have shown better performance over conventional fixed-time and rule-based methods. However, 
these RL-based controllers rely heavily on interactions with the environment and require significant amounts of training data, rendering them unsuitable for on-device computation. 
Researchers have attempted to address such sample efficiency issues by adopting meta-learning \citep{zang2020metalight, zhu2023metavim} or multi-task learning \citep{zhu2022mtlight}. This involves pre-training a well-initialized or well-generalized controller on source tasks and deploying the controller to target tasks with zero or few training steps. 
Model-based RL, which learns the dynamics model of the environment and generates imaginary training data for policy learning, is also adopted to improve sample efficiency~\citep{huang2021modellight, devailly2022model}.

Although these methods can help improve sample efficiency, they remain challenging for real-world applications since the undertrained controller at the early training or pre-training phase may cause serious congestion. Offline RL~\citep{prudencio2023survey}, which learns policy on historical offline datasets and requires no interaction with the environment during pre-training, has proven to be a promising technique for solving real-world problems, e.g., automated driving~\citep{diehl2023uncertainty}, and robotics~\citep{gurtler2023benchmarking}. 
Offline RL has also achieved primary success in TSC \citep{dai2021traffic, kunjir2022offline} while limited to single-intersection control and failing to match the performance of online RL-based methods.

\begin{figure}[ht]
\centering
\includegraphics[width=0.65\columnwidth]{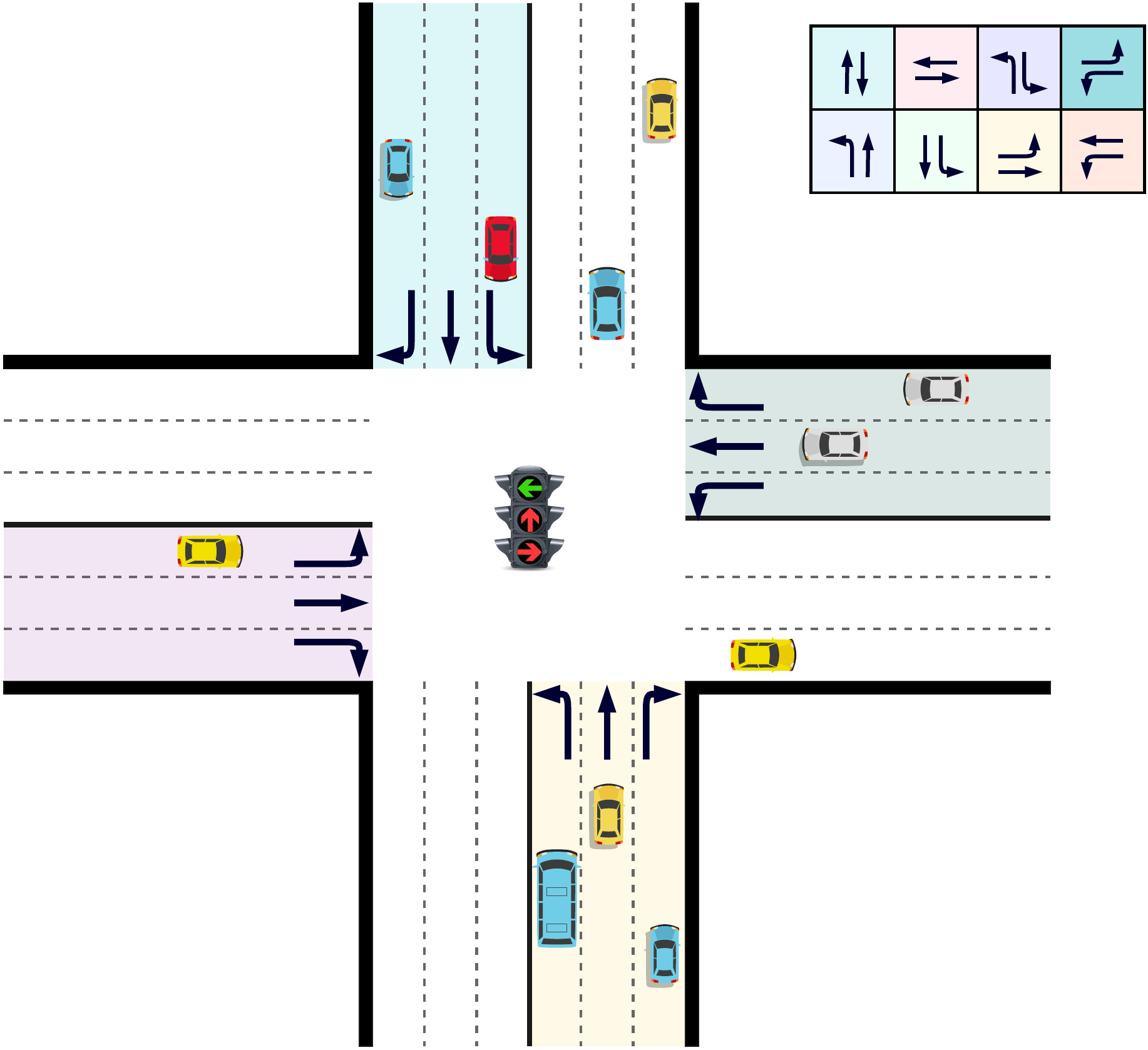} 
\caption{The standard signalized intersection with 12 incoming lanes. 8 different phases are shown on the top right.}
\label{fig1}
\end{figure}

Decision Transformer (DT)~\citep{chen2021decision}, an offline RL method that formulates policy learning as a sequence modeling problem and employs Transformers to solve it, exhibits superior performance on several offline RL tasks. Given that traffic signal control involves understanding the patterns in the flow of traffic over time and can be naturally considered a sequence modeling problem, in this work, we leverage DT to solve TSC issues in both single- and multi-intersection TSC scenarios. We introduce a novel lightweight DT-based algorithm, named DTLight. Specifically, we first build offline datasets on eight TSC scenarios using both a conventional and an RL-based TSC method.
To reduce the model size of DTLight and make it applicable, we pre-train a teacher DTLight in large size and use the knowledge distillation technique \citep{sanh2019distilbert} to learn a smaller-sized student DTLight that retains comparable abilities. To make DTLight adaptable online, we introduce a COMPACTER++ adapter module \citep{karimi2021compacter} to each layer of the Transformer structure of DTLight. This implies that we only need to update the adapter module during online adaptation, thus avoiding full fine-tuning of the entire model. Further enhancements are applied to make DTLight aligned with real-world TSC settings.

We summarize the contributions of this work:
\textbf{(i)} We create DTRL, the first publicly available TSC offline datasets on eight TSC scenarios, including both real-world and synthetic scenarios with single and multiple intersections; 
\textbf{(ii)} To the best of our knowledge, this is the first attempt to employ offline-to-online reinforcement learning in addressing both single and multiple signalized intersections. Specifically, we propose DTLight, a novel lightweight traffic signal controller based on Decision Transformer. By adopting knowledge distillation, we reduce the model size of the pre-trained model for feasible on-device computation. Additionally, we employ adapter modules for online adaptation extending beyond transfer learning. Our DTLight is further enhanced to be more applicable to real-world TSC problems;
\textbf{(iii)} Simple yet powerful, DTLight exhibits high sample efficiency, low on-device computation cost, and significantly improved performance over both conventional and offline/online RL-based TSC methods through rigorous experiments. 
Our datasets and code are openly available.

\begin{figure*}[ht]
\begin{center}
\includegraphics[width=0.9\textwidth]{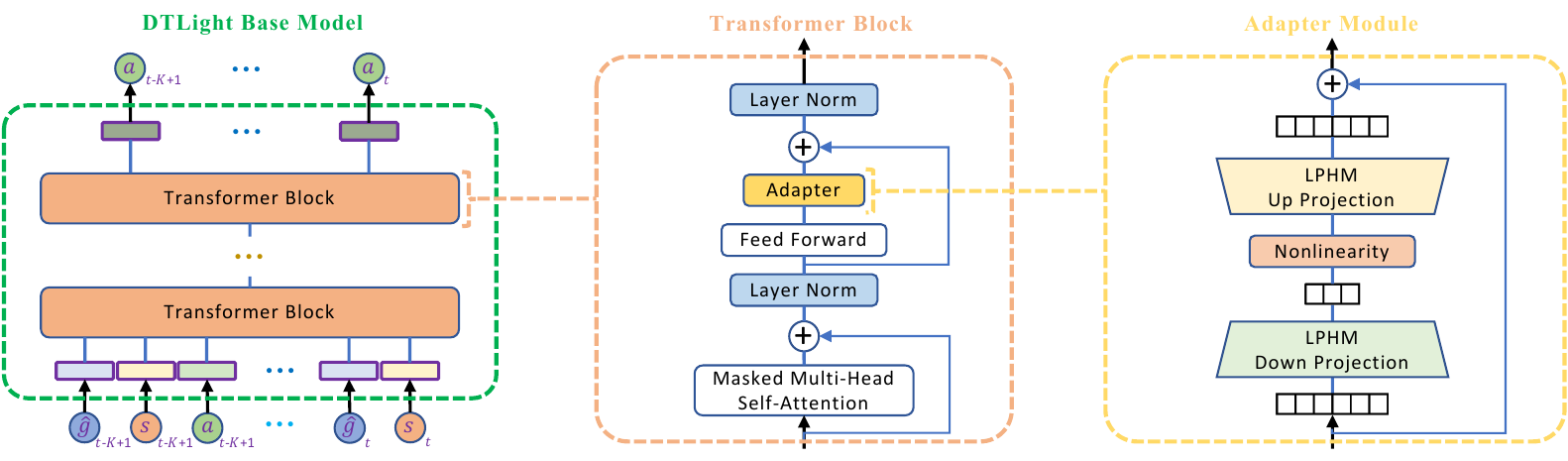}
\end{center}
\caption{Model structure of DTLight. \textbf{Left:} DTLight takes in recent context and outputs actions. \textbf{Middle:} We add adapter modules only after each feed-forward layer. \textbf{Right:} The COMPACTER++ adapter module consists of low-rank parameterized hypercomplex multiplication (LPHM) down and up projection layers, and a nonlinearity function.}
\label{fig2}
\end{figure*}

\section{Related Work}
\textbf{Reinforcement Learning-Based Traffic Signal Control.}
The traffic signal control problem can be formulated as a Markov decision process and solved with reinforcement learning. Prior works have delved into the state representation and reward function of TSC.
For instance, Advanced-XLight \citep{zhang2022expression} proposes an advanced traffic state to combine efficient pressure \citep{chen2020toward} with running vehicles. 
In addition to controlling each signalized intersection with a separate controller, researchers tried to consider multiple intersections in a network as a fully cooperative system and solve it with multi-agent RL algorithms. 
RESCO \citep{ault2021reinforcement} introduces an RL benchmark for both single- and multi-intersection TSC.
MABRL \citep{zhu2023multi} employs a broad network to handle global information and devise a dynamic interaction mechanism to facilitate coordination between agents.
Some researchers focus on solving different TSC scenarios with a well-initialized or well-generalized controller. MetaVIM \citep{zhu2023metavim} and MTLight \citep{zhu2022mtlight} respectively propose meta-learning and multi-task learning pipelines. 
Offline RL has also been studied in TSC. 
\citet{kunjir2022offline} applies a model-based offline RL algorithm DAC-MDP to solve a single roundabout control scenario.

\textbf{Transformers in Reinforcement Learning.}
The Transformer architecture \citep{vaswani2017attention} has shown success in a variety of supervised learning tasks in recent years due to its strong sequence modeling capability \citep{li2023survey}. Applying Transformer to RL has also attracted the attention of some researchers. 
Transformer architecture is adopted for representation learning to model local per-timestep sequences \citep{tang2021sensory} and temporal sequences \citep{melo2022transformers}. Additionally, 
Transformers form the backbone of the world model in some model-based RL algorithms. IRIS \cite{micheli2022transformers} learns the world model with a Transformer backbone via autoregressive learning and achieves state-of-the-art performance on the Atari 100k games. Recently, some works treat learning a policy in RL as a sequence modeling problem. Decision Transformer \citep{chen2021decision} exploits Transformer as a sequential decision-making model to solve offline RL and achieves comparable performance as conventional offline RL methods. Online DT (ODT) \citep{zheng2022online} further enhances the online performance of DT by fine-tuning a model pre-trained offline.

\renewcommand{\algorithmicrequire}{\textbf{Require:}}
\renewcommand{\algorithmicensure}{\textbf{Output:}}
\begin{algorithm}[ht]
\caption{DTLight for Traffic Signal Control} 
\label{a1} 
\begin{algorithmic}[1]
    \Require offline dataset $D_{offline}$, number of teacher training iterations $J$, distillation iterations $M$, and online fine-tuning iterations $W$
    \Statex \textbf{\textit{Pre-training Teacher Policy $\pi_{\theta_{t}}$}:}
    \State Initialize $\pi_{\theta_{t}}$
    \For{$j$ in $J$}
        \State Sample $X$ batches of sub-trajectories from $D_{offline}$  
        \For{each batch of sub-trajectories $ \left \{(\mathbf{s}, \mathbf{a}, \hat{\mathbf{g}}) \right \}$ }
            \State \parbox[t]{0.86\linewidth}{Update $\theta_{t}$ by minimizing $L_{dt}(\theta_{t})$ based on $ \left \{(\mathbf{s}, \mathbf{a}, \hat{\mathbf{g}}) \right \}$}
        \EndFor
    \EndFor
    \Statex \textbf{\textit{Knowledge Distillation}:}
    \State Initialize student policy $\pi_{\theta_{s}}$
    \For{$m$ in $M$}
        \State Sample $Y$ batches of sub-trajectories from $D_{offline}$
        \For{each batch of sub-trajectories $ \left \{(\mathbf{s}, \mathbf{a}, \hat{\mathbf{g}}) \right \}$ }
            \State \parbox[t]{0.86\linewidth}{Update $\theta_{s}$ by minimizing $L(\theta_{s})$ based on $\pi_{\theta_{t}}$ and $ \left \{(\mathbf{s}, \mathbf{a}, \hat{\mathbf{g}}) \right \}$}
        \EndFor
    \EndFor
    \Statex \textbf{\textit{Online Fine-tuning}:}
    \State Online dataset $D_{online} \leftarrow$ {top $C$ trajectories in $D_{offline}$}
    \For{$w$ in $W$}
        \State \parbox[t]{0.93\linewidth}{Rollout trajectory $\tau$ from the environment using policy $\pi_{\theta_{s}}$}
        \State $D_{online} \leftarrow \left\{D_{online} \backslash\{\text{min-return trajectory}\}\right\} \bigcup\{\tau\}$
        \State Label $\hat{g}$ of $\tau$ using true rewards
        \State Sample $Z$ batches of sub-trajectories from $D_{online}$
        \For{each batch of sub-trajectories $ \left \{(\mathbf{s}, \mathbf{a}, \hat{\mathbf{g}}) \right \}$ }
            \State \parbox[t]{0.86\linewidth}{Update parameters of adapter modules and output layers in $\pi_{\theta_{s}}$ while fixing other parameters}
        \EndFor
    \EndFor
\end{algorithmic}
\end{algorithm}

\section{Preliminaries}

\subsection{Traffic Signal Control Problem}
To optimize traffic flow and improve safety, traffic signal controllers should accurately capture the characteristics of each intersection, including approaches, lanes, traffic flow, and phases, as shown in Figure~\ref{fig1}.

\textbf{Approaches and Lanes.} 
An approach is a section of a road with one or more lanes that leads to a junction. Various lane kinds correspond to various vehicle movements, including left turn, right turn, and through movement. 

\textbf{Traffic Flow.} The movement of vehicles through an intersection or a road network is referred to as traffic flow in the context of TSC. TSC aims to control traffic flow for effective and secure vehicle movement.

\textbf{Phases.} A phase is a grouping of various traffic lights that are activated simultaneously in various lanes, regulating the orderly passage of vehicles at the intersection and avoiding collisions. There are eight distinct phases in this particular crossing, as shown in Figure\ref{fig1}. Since it typically only imposes minor restrictions, the right-hand turn is not included in every phase.

\subsection{Offline Reinforcement Learning and Decision Transformer}
Reinforcement learning aims to solve Markov Decision Process described by the tuple $(\mathcal{S}, \mathcal{A}, \mathcal{P}, \mathcal{R}, \gamma)$ \citep{sutton2018reinforcement}. The MDP tuple consists of state $s \in \mathcal{S}$, action $a \in \mathcal{A}$, state transition $\mathcal{P}(s'|s, a)$, reward function $r = \mathcal{R}(s, a)$, and discount factor $\gamma$ ~\citep{noaeen2022reinforcement}. The objective of reinforcement learning is to learn a policy $\pi$ that maximizes the expected future discounted return $E_{\pi}\left[\sum_{t=0}^{\infty} \gamma^{t} \mathcal{R}\left(s_{t}, a_{t}\right)\right]$, where $s_{t}$ and $a_{t}$ represent state and action at time $t$, respectively. Offline RL aims to learn a policy without interactions with the environment but purely from limited historical experience, i.e., an offline dataset typically composed of tuples of the form $(s, a, r, s')$ and additional information.
While offline RL can be more cost-effective and safer, it is more difficult as it avoids real-world experimentation and is unable to receive any feedback.

Decision Transformer~\citep{chen2021decision} leverages the Transformer architecture~\citep{vaswani2017attention}, specifically Generative Pre-trained Transformer (GPT)-like architecture \citep{radford2018improving, radford2019language}, to solve offline RL problems by autoregressive modeling trajectories.
DT generates action autoregressively given the recent context, a $K$-step sub-trajectory $\tau$ including 3 types of tokens: states, return-to-go (RTG) $\hat{g}_{t} = \sum_{t'=t}^{T}r_{t'}$, and actions \citep{chen2021decision}.
\begin{equation}
    \tau=\left(s_{t-K+1}, a_{t-K+1}, \hat{g}_{t-K+1}, \ldots, s_{t}, a_{t}, \hat{g}_{t}\right).
\end{equation}
Equivalently, DT learns a deterministic policy $\pi(a_{t}|\textbf{s}_{-K, t}, \hat{\textbf{g}}_{-K, t})$, where $\textbf{s}_{-K, t}$ and $\hat{\textbf{\textbf{g}}}_{-K, t}$ denote the latest $K$-step states and RTG, respectively. Note that the latest actions are generated autoregressively and fed into the policy when predicting the next action; hence, they are omitted in the policy.

\section{Methodology}

\label{Methodology}
In this section, we first define the Markov Decision Process formulation of traffic signal control. 
Then we introduce the structure of the DTLight base model, followed by knowledge distillation and online fine-tuning.
Subsequently, we present how to extend the proposed solution to multi-agent settings and summarize the implementation details.
Algorithm~\ref{a1} illustrates the three main learning processes of DTLight: pre-training a teacher DTLight base model, knowledge distillation which learns a smaller-size student DTLight model, and online fine-tuning by merely updating the adapter modules. The model architecture of DTLight is depicted in Figure \ref{fig2}. 

\subsection{Markov Decision Process Formulation}
\label{mdp}
We formulate TSC as an MDP and the state, action, and reward function are defined as follows.

\textbf{State.} 
For the DTLight agent, its state includes the queue length, number of approaching vehicles, and accumulated waiting time of stopped vehicles at the corresponding signalized intersection. 

\textbf{Action.} 
Typically, the choice of the phase for an intersection at the next time step is defined as the action $a_{t}$, which controls the vehicle movement \citep{ault2021reinforcement}.

\textbf{Reward.} 
We adopt the commonly-used pressure $p$ \citep{chen2020toward} as the reward function, i.e., $r = -p$. 
Specifically, pressure $p_{t}$ is the difference between the sum of queue lengths of all entering lanes and those of exiting lanes at a signalized intersection at time step $t$.
It is easy to implement and is relevant to boosting traffic efficiency \citep{ault2021reinforcement}.

\subsection{DTLight Base Model} 
\label{dt}
Given traffic’s sequential and time-dependent nature, understanding temporal relationships is critical for efficient signal control. DT \citep{chen2021decision}, built with self-attention layers and residual connections, excels in sequence modeling by leveraging historical data to predict future traffic conditions and optimize signal timing.
Therefore, we adopt DT as our base model (as shown in the left part of Figure~\ref{fig2}).
Different from DT settings, we propose to learn a stochastic policy that models the discrete action distribution with a categorical distribution. This enables exploration when applying DTLight to online adaptation.
\begin{equation}
    \pi_{\theta}(a_{t}=a^{i}|\textbf{s}_{-K, t}, \hat{\textbf{g}}_{-K, t}) = p_{i},
\end{equation}
\noindent where $a^{i}$ denotes the $i$-th action in the discrete action space of size $N$ and $\sum_{i=1}^{N}p_{i} = 1$. We use $(\mathbf{a}, \mathbf{s}, \hat{\mathbf{g}}) \sim \mathcal{T}$ to represent sampling from data distribution $\mathcal{T}$.
The stochastic policy $\pi_{\theta}$ with learnable parameters $\theta$ is learned by minimizing the negative log-likelihood loss over the training trajectories.
\begin{equation}
    \begin{array}{c}
        l(\theta)=\frac{1}{K} \mathbb{E}_{(\mathbf{a}, \mathbf{s}, \hat{\mathbf{g}}) \sim \mathcal{T}}\left[-\log \pi_{\theta}(\mathbf{a} \mid \mathbf{s}, \hat{\mathbf{g}})\right] \\
        =\frac{1}{K} \mathbb{E}_{(\mathbf{a}, \mathbf{s}, \hat{\mathbf{g}}) \sim \mathcal{T}}\left[-\sum_{t=1}^{K} \log \pi_{\theta}\left(a_{t} \mid \mathbf{s}_{-K, t}, \hat{\mathbf{g}}_{-K, t}\right)\right].
    \end{array}
\end{equation}
Following Online DT~\citep{zheng2022online}, we also add an entropy loss to further encourage exploration.
\begin{equation}
    \begin{array}{c}
        H_{\theta}^{\mathcal{T}}[\mathbf{a} \mid \mathbf{s}, \hat{\mathbf{g}}]=\frac{1}{K} \mathbb{E}_{(\mathbf{s}, \hat{\mathbf{g}}) \sim \mathcal{T}}\left[H\left[\pi_{\theta}(\mathbf{a} \mid \mathbf{s}, \hat{\mathbf{g}})\right]\right] \\
        =\frac{1}{K} \mathbb{E}_{(\mathbf{s}, \hat{\mathbf{g}}) \sim \mathcal{T}}\left[\sum_{k=1}^{K} H\left[\pi_{\theta}\left(a_{k} \mid \mathbf{s}_{-K, k}, \hat{\mathbf{g}}_{-K, k}\right)\right]\right],
    \end{array}
\end{equation}
\noindent where $H[\cdot]$ denotes the Shannon entropy \citep{zheng2022online}. Therefore, the training loss of DTLight becomes
\begin{align}
\label{equ5}
    L_{dt}(\theta) &= \mathbb{E}_{(\mathbf{s}, \mathbf{a}, \hat{\mathbf{g}}) \sim \mathcal{T}}\left[-\log \pi_{\theta}(\mathbf{a} \mid \mathbf{s}, \hat{\mathbf{g}})\right] \nonumber \\
    &\quad - \lambda \mathbb{E}_{(\mathbf{s}, \hat{\mathbf{g}}) \sim \mathcal{T}}\left[H\left[\pi_{\theta}(\cdot \mid \mathbf{s}, \hat{\mathbf{g}})\right]\right],
\end{align}
\noindent where $\lambda$ is a learnable temperature parameter used for regularization.

\subsection{Knowledge Distillation}
\label{kd}

There is always a gap between large models showing outstanding performance and affordable models in real applications. This also applies to TSC when deploying RL-based methods to solve real-world problems. We alleviate this issue by leveraging knowledge distillation, that is, a student model with a reduced model size is learned by transferring knowledge from a large and cumbersome teacher model without compromising the performance \citep{hinton2015distilling}. This works by making the student's predictions match the soft targets generated by the strong teacher model.

In this work, we first learn a large and stronger teacher DTLight model by minimizing the $L_{dt}$ introduced in Equation \ref{equ5} over the offline dataset (line 1-7 of Algorithm \ref{a1}). Subsequently, a smaller student DTLight model is trained (line 8-14 of Algorithm \ref{a1}) with a loss function linearly combining a distillation loss (soft loss) $L_{kd}$ and a hard loss $L_{dt}$.
\begin{equation}
    L = \alpha L_{kd} + \beta L_{dt},
\end{equation}
\begin{equation}
    L_{kd} = -\sum_{i}^{N} t_{i} \log \left(s_{i}\right),
\end{equation}
\noindent where $\alpha$ and $\beta$ are loss weights. $t_{i}=\frac{\exp \left(z_{i} / T\right)}{\sum_{j}^{N} \exp \left(z_{j} / T\right)}$ and $s_{i}=\frac{\exp \left(v_{i} / T\right)}{\sum_{j}^{N} \exp \left(v_{j} / T\right)}$ denote the generated action distributions by teacher and student, respectively. Further, $z_i$ is the output logits for action $i$, and $T$ is a temperature factor controlling the smoothness of the generated action distribution.

\subsection{Online Fine-tuning with Adapter Modules}
\label{am}
Even though we can learn a relatively smaller DTLight model with knowledge distillation, the backbone Transformer model is still hard to finetune in low-budget TSC systems when online adaptation is required. To make the proposed solution more applicable in the real world, we introduce an adapter module, small learnable neural networks added to specific layers in the pre-trained model \citep{houlsby2019parameter}, to eliminate the need for retraining the entire model of DTLight. 
As illustrated in the middle of Figure~\ref{fig2}, during fine-tuning, the layer normalizations and lightweight adapter modules with a small number of additional parameters are retrained to adapt to the online environment while keeping the pre-trained parameters of the large Transformer model fixed.

We adopt COMPACTER++ \cite{karimi2021compacter} in our work given that COMPACTER++ has a complexity of $\mathcal{O}(k+d)$ compared with a complexity of $\mathcal{O}(kd)$ for vanilla Adapter \citep{houlsby2019parameter}. 
COMPACTER++ keeps only the adapter after the feed-forward layer while removing the adapter module after the self-attention layer in each Transformer block. The right part of Figure~\ref{fig2} illustrates the architecture of the COMPACTER++ adapter module. It replaces the feed-forward down and up projection layers of the vanilla Adapter with low-rank parameterized hypercomplex multiplication (LPHM) layers. The resulting COMPACTER++ adapter module can be formulated as:
\begin{equation}
A(\boldsymbol{x})=\operatorname{LPHM}^{U}\left(\operatorname{GeLU}\left(\operatorname{LPHM}^{D}(\boldsymbol{x})\right)\right)+\boldsymbol{x},
\end{equation}
\noindent where $\operatorname{GeLU}$ is the Gaussian error linear units (nonlinearity) and $x$ is the input hidden state. $U$ and $D$ denote up and down projections, respectively.

When augmenting trajectories online, instead of manually setting the initial RTG values as hyperparameters, we initialize them as the product of the maximum return values in offline trajectories and a scale factor $\gamma_{online}$. This better exploits the offline data and makes implementation for controlling multiple intersections easier, since manually setting the RTG value for each agent is cumbersome. Similarly, we set an evaluation RTG scale factor $\gamma_{eval}$ for policy evaluation.
Following some other online fine-tuning settings as in online decision transformer \citep{zheng2022online}, we initialize the replay buffer (online dataset in Algorithm \ref{a1}) by choosing the trajectories with the top values of return in the offline dataset. Instead of storing transitions in the replay buffer, we replace the old trajectory of the minimum return value with the whole new trajectory obtained by interacting with the environment using the student policy (line 18 of Algorithm \ref{a1}). The RTG values of the added new trajectories are labeled using true rewards: $ \hat{\mathbf{g}}_{t}=\sum_{t'=t}^{|\tau|} r_{t'}, 1 \leq t \leq |\tau| $.

\begin{table*}
\footnotesize
\centering
\caption{
Average delay performance in seconds (mean$\pm$standard deviation) of different methods on eight TSC scenarios over five random seeds. \textit{Improvements 1} and \textit{2} denote improvements of DTLight-o over the best baseline and behavior policy, respectively. The bold values indicate the highest performance among all methods, excluding DTLight-t models, which are not considered for direct applications in this study due to the large size.
} 
\label{T1}
\begin{tabular}{p{2.9cm}|p{1.2cm}p{1.2cm}|p{1.2cm}p{1.2cm}|p{1.2cm}p{1.5cm}|p{1.5cm}p{1.8cm}}
\toprule
  & \multicolumn{2}{c|}{\textbf{Real Single}} & \multicolumn{2}{c|}{\textbf{Synthetic Single}} & \multicolumn{2}{c|}{\textbf{Real Multiple}} & \multicolumn{2}{c}{\textbf{Synthetic Multiple}} \\
  & Cologne & Ingolstadt & 3-Lane & 2-Lane & 3-Inter & 8-Inter & Grid 4 × 4 & Avenues 4 × 4 \\
\hline
Fixed Time & 54.5 & 38.1 &  24.7 &  86.7 & 37.5 & 58.0 & 81.7 & 1256.6 \\
MaxPressure &  27.6 & 22.6 &  16.5 & 35.1 & 23.4 & 28.0 & 50.7 & 825.6 \\
IDQN &  26.0$\pm$0.5 & 21.4$\pm$0.5 &  16.6$\pm$0.4 &  36.7$\pm$6.8 & 23.9$\pm$1.1 & 22.0$\pm$0.3 & 32.9$\pm$0.2& 1168.3$\pm$194.4\\
IPPO  &  43.2$\pm$11.8 & 20.9$\pm$0.2 &  38.4$\pm$4.0 &  71.9$\pm$8.2& 24.0$\pm$0.4 & 21.6$\pm$0.1 & 42.9$\pm$1.0& 706.5$\pm$19.0 \\
MPLight  &  28.7$\pm$1.6 & 28.3$\pm$0.8 &  16.2$\pm$0.2 &  31.0$\pm$0.2 & 83.6$\pm$27.9 & 60.4$\pm$20.1 & 46.9$\pm$0.9 & 837.7$\pm$96.4 \\
FMA2C  &  30.7$\pm$0.3 & 27.0$\pm$1.5 & 17.3$\pm$0.9 & 34.7$\pm$3.1 & 26.8$\pm$0.1 & 33.8$\pm$0.4 & 95.2$\pm$1.9 & 673.5$\pm$3.1 \\

\hline
EMP (Behavior)  & 29.7   & 21.2     & 16.1 & 33.3 & 23.5 & 28.2   & 48.39 & 646.2  \\
DTLight-t (EMP)  &  24.4$\pm$0.1 & 20.7$\pm$0.3 & 13.3$\pm$0.2 & 18.5$\pm$0.2 & 17.0$\pm$0.1 & 32.2$\pm$11.1 & 175.6$\pm$30.5 & 803.9$\pm$2.8 \\
DTLight-s (EMP)  &  24.6$\pm$0.4 & 20.7$\pm$0.2 & 15.8$\pm$0.3 & 19.1$\pm$0.5 & 18.9$\pm$0.3 & 66.1$\pm$39.3 & 446.8$\pm$128.0 & 1004.9$\pm$335.2 \\
DTLight-o (EMP)  &  \textbf{24.2}$\pm$0.3 & 20.4$\pm$0.2 & 13.3$\pm$0.2 & 18.0$\pm$0.3 & 17.6$\pm$0.4 & 23.0$\pm$0.1 & 49.6$\pm$0.4 & \textbf{574.3}$\pm$97.8 \\

\hline
\textit{Improvements 1} (\%) &  \textit{7.0} & \textit{2.2}  & \textit{17.6}  & \textit{41.9} & \textit{26.5}& - & -& \textit{11.1} \\
\textit{Improvements 2} (\%) &  \textit{18.7} & \textit{3.8}  & \textit{17.5}  & \textit{45.9} & \textit{25.1} & \textit{18.8} & - & \textit{11.1} \\

\hline
IDQN(Behavior)             & 28.2             & 19.9     & 16.0  & 30.9 & 21.6                      & 22.0               & 37.81 & 778.7                \\
DTLight-t (IDQN)  &  23.8$\pm$0.2 & 19.3$\pm$0.9 & 12.2$\pm$0.3 & 17.7$\pm$0.2 & 16.6$\pm$0.1 & 18.2$\pm$0.1 & 38.3$\pm$8.4 & 782.7$\pm$7.3 \\
DTLight-s (IDQN)  &  40.5$\pm$17.8 & 19.4$\pm$0.2 & 14.2$\pm$0.3 & 18.7$\pm$0.1 & 16.9$\pm$0.6 & 19.6$\pm$1.6 & 61.2$\pm$27.3 & 777.9$\pm$12.6 \\
DTLight-o (IDQN)  &  24.4$\pm$0.6 & \textbf{19.0}$\pm$0.5 & \textbf{12.0}$\pm$0.1 & \textbf{17.8}$\pm$0.4 & \textbf{16.0}$\pm$0.4 & \textbf{18.3}$\pm$0.2 & \textbf{26.7}$\pm$0.1 & 752.3$\pm$6.3 \\

\hline
\textit{Improvements 1} (\%) & \textit{7.0} &\textit{9.2}  & \textit{25.5}  & \textit{42.6} & \textit{31.6} & \textit{15.2} & \textit{18.94} & -\\
\textit{Improvements 2} (\%) & \textit{13.7} & \textit{4.7} & \textit{24.8}  & \textit{42.4} & \textit{25.9} & \textit{16.8} & \textit{29.36} & \textit{3.2} \\

\bottomrule
\end{tabular}
\end{table*}

\subsection{Multi-Agent Settings}
To address traffic signal control of multiple intersections in a road network, following decentralized settings~\cite{chu2019multi}, we control each signalized intersection using an independent DTLight agent. This is a simple but effective approach to addressing the combinatorially large joint action space. However, it also suffers from partial observability, i.e., each agent can only access its local state space. While global state sharing is difficult in the real-world traffic system, similar to MA2C~\cite{chu2019multi, ault2021reinforcement}, we introduce neighborhood information to each agent to enhance observability and facilitate coordination between different agents. Each agent not only observes its local state but also shares information about its neighbors, which is feasible under limited communication. Given that traffic conditions in the neighborhood, intersections have no direct impact on the current state of the target intersection, we adopt a discount factor $\delta$ to scale down the states from neighbors. Therefore, the state of each independent agent is the combination of the local state and discounted neighborhood states, i.e.,
$
s = \left[s_{i}\right] \cup \delta\left[s_{j}\right]_{j \in \mathcal{N}_{i}}
$,
where $\mathcal{N}_{i}$ denotes the neighborhood intersections. In multi-intersection scenarios, DTLight integrates both local and neighborhood information through its attention mechanism, allowing it to capture complex dependencies among various interactions.

\subsection{Implementation Details} 
\label{ai}

For easy reproduction of our work, we present the implementation details of DTLight in this part. 
We use GPT2 architecture \citep{radford2019language} as the backbone model to learn both the teacher and student policies of DTLight. The teacher model incorporates 6 layers and 8 attention heads, with an embedding dimension of 512, while the student model uses 2 layers and 2 attention heads, with an embedding dimension of 256. The initial temperature $\lambda$ for entropy loss and Softmax temperature $T$ for knowledge distillation are 0.1 and 8, respectively. When combining the distillation loss and hard loss, $\alpha$ is 0.4, and $\beta$ is 1.
The initial RTG values for evaluation and online data augmentation are set to the product of the maximum return values in offline trajectories and scale factors which are 0.2 and 0.3 for evaluation and online fine-tuning, respectively. 
We designate a batch size of 256 and 
pre-train the teacher and student models with 2000 and 3000 gradient updates, respectively. DTLight is fine-tuned online for 10 episodes, applying 300 gradient updates per episode.
Most of the hyperparameters are selected through the random search, an efficient hyperparameter optimization method \citep{goodfellow2016deep}.

\begin{table}
\footnotesize
\centering
\caption{
Model size and training time comparison. 
}
\label{T2}
\begin{tabular}{l|ll|l}
\toprule 
 & \multicolumn{2}{c|}{\textbf{Model Sizes}} &  Training\\
 & \# param. (M)      & Percentage (\%)     & Time (s) \\ 
\hline
DTLight-t & 19.44  & 100 & 520 \\
DTLight-s & 1.84  & 9.47 & 326  \\
Adapter   & 0.002  & 0.008 & 35  \\
\bottomrule
\end{tabular}
\end{table}

\section{Experiments}
\label{Experiments}

We conduct experiments using a commonly used simulator, SUMO \cite{lopez2018microscopic} on an Intel(R) Core(TM) i9-12900K CPU and a single NVIDIA GEFORCE RTX 3090 GPU. DTLight is implemented based on the open-source code of RESCO \citep{ault2021reinforcement} and the Transformers library~\cite{wolf-etal-2020-transformers}. 

\textbf{DTRL: Datasets for Traffic Signal Control with Offline Reinforcement Learning.}
To facilitate research on TSC with offline reinforcement learning, in this work, we create DTRL, which includes 16 offline datasets. These datasets are generated by two different behavior policies: a conventional TSC method EMP (Max Pressure~\citep{chen2020toward} with additional $\epsilon$-greedy exploration) and an RL-based TSC method IDQN \citep{ault2021reinforcement}. We implement each behavior policy on eight TSC scenarios for 100 (training) episodes and store the generated trajectories in JSON files (offline datasets).
Specifically, we create two synthetic single-intersection control scenarios, i.e., a signalized intersection with four three-lane approaches named \textit{3-Lane} and an intersection with four two-lane approaches named \textit{2-Lane}. For these two scenarios, we simulate different traffic conditions, ranging from light to heavy. Three different arrival rates are used for each scenario, i.e., 0.25, 0.5, and 1 $\text{vehicles}/s$ within a time period of 3600 seconds for 3-Lane, and 0.2, 0.4, and 0.8 $\text{vehicles}/s$ for 2-Lane due to the reduced number of lanes. We adopt the other six scenarios from RESCO \citep{ault2021reinforcement} benchmark: two real-world single-intersection control scenarios from two cities, \textit{Cologne} \citep{beckmann2006integrierte} and \textit{Ingolstadt} \citep{lobo2020intas}, two real-world multi-intersection control scenarios with 3 and 8 signalized intersections (\textit{3-Inter} and \textit{8-Inter} for short), respectively \citep{beckmann2006integrierte}, and two synthetic 4 × 4 symmetric network scenarios, i.e., \textit{Grid 4 × 4} \cite{chen2020toward} and \textit{Avenues 4 × 4} \cite{ma2020feudal}. The task horizon of each TSC scenario is 3600 seconds, with control steps occurring every 10 seconds, except for Avenues 4 × 4, where the control step is 5 seconds. 

\textbf{Baselines.}
We use both conventional TSC methods and online RL-based methods as baselines for comparison:
(1) \textbf{Fixed Time} operates on a predefined cycle and duration;
(2) \textbf{Max Pressure} \citep{chen2020toward} chooses joint phases with the maximum pressure;
(3) \textbf{EMP} ($\epsilon$-greedy Max Pressure);
(4) \textbf{IDQN} 
(independent DQN) 
\citep{ault2019learning} hires separate DQN agents for with a negative waiting time;
(5) \textbf{IPPO} 
(independent proximal policy optimization) 
\citep{ault2021reinforcement} employs numerous PPO agents with network structures identical to IDQN;
(6) \textbf{MPLight} \citep{chen2020toward} shares parameters across all DQN agents and rewards them based on pressure;
(7) \textbf{FMA2C} 
(Feudal Multi-agent Advantage Actor-Critic)
~\citep{ault2021reinforcement, ma2020feudal}, a variant of MA2C \citep{chu2019multi}, uses A2C agents to control each intersection and suggests improving coordination between agents through neighborhood information, and discounted rewards and states.
We adopt the official implementation and results of the above-mentioned baselines from RESCO benchmark \citep{ault2021reinforcement}.

\subsection{Results Analysis}

We compare the performance of DTLight with all the baselines on the above-mentioned eight scenarios.
Particularly, we denote the teacher DTLight and student DTLight pre-trained on offline datasets generated by EMP or IDQN by \textit{DTLight-t (EMP)} and \textit{DTLight-s (EMP)}, or \textit{DTLight-t (IDQN)} and \textit{DTLight-s (IDQN)}, respectively. \textit{DTLight-o} denotes online fine-tuned DTLight. 

\textbf{Performance comparison.}
Since traffic congestion can be reflected by the signal-induced dealy of vehicles, we present the average delay of different TSC methods in Table \ref{T1}, which follows a similar pattern as other evaluation metrics, e.g., average travel time~\cite{ault2021reinforcement}. 
Lower values indicate better performance.
Experiments show that IDQN and MPLight demonstrate convergence after 100 online episodes, whereas IPPO and FMA2C require 1400 episodes to achieve convergence. Notably, we only finetune DTLight with 10 online episodes, which significantly reduces the reliance on online interactions with the environment.

As illustrated in Table~\ref{T1}, when trained on offline datasets generated by EMP, DTLight-o achieves superior performance over all baselines and the behavior policy (EMP) on 6 and 7 scenarios, respectively. DTLight-s trained purely offline also outperforms all baselines on 5 out of 8 scenarios. For scenarios where DTLight-s fails to learn qualified policies, DTLight-o restores its high performance via a small number of interactions with the environment. This might be due to that EMP lacks the ability to explore multi-intersection scenarios, resulting in low-quality offline datasets.
When pre-trained on datasets generated by IDQN, DTLight-s and DTLight-o outperform all baselines on 6 and 7 out of 8 scenarios, yielding improvements up to 39.7\% and 42.6\%, respectively. DTLight-s and DTLight-o surpass the performance of the behavior policy (IDQN) on all scenarios except for DTLight-s on Grid 4 × 4.
The close resemblance of results between DTLight-t and DTLight-s demonstrates the effectiveness of our knowledge distillation approach in reducing model complexity while preserving its capability.
The observed performance improvement of DTLight-o indicates that online fine-tuning of adapters can effectively and significantly reduce model training parameters and iterations while further enhancing model performance.

\textbf{Size and training speed.}
In order to make computation feasible in a real deployment, we employ knowledge distillation and adapter modules to reduce the model size and training time of DTLight.
As illustrated in Table~\ref{T2}, DTLight-t contains 19.44 million training parameters and costs 520$s$ training time 
using a single NVIDIA GEFORCE RTX 3090 GPU 
for a single agent to get a competitive pre-trained policy. However, knowledge distillation reduces the model size to 1.84 million parameters (only 9.47\% of parameters of the teacher model) and offline training time to 326$s$ on the same device to achieve comparable performance. The adapter module further reduces the online training parameters to 0.002 million and training time to 35$s$, which makes it more applicable to fine-tune DTLight online on a real TSC system.

\subsection{Ablation Study}

We verify the importance of some key components by varying the structure of DTLight.

\begin{figure}
\begin{center}
\includegraphics[width=0.47\textwidth]{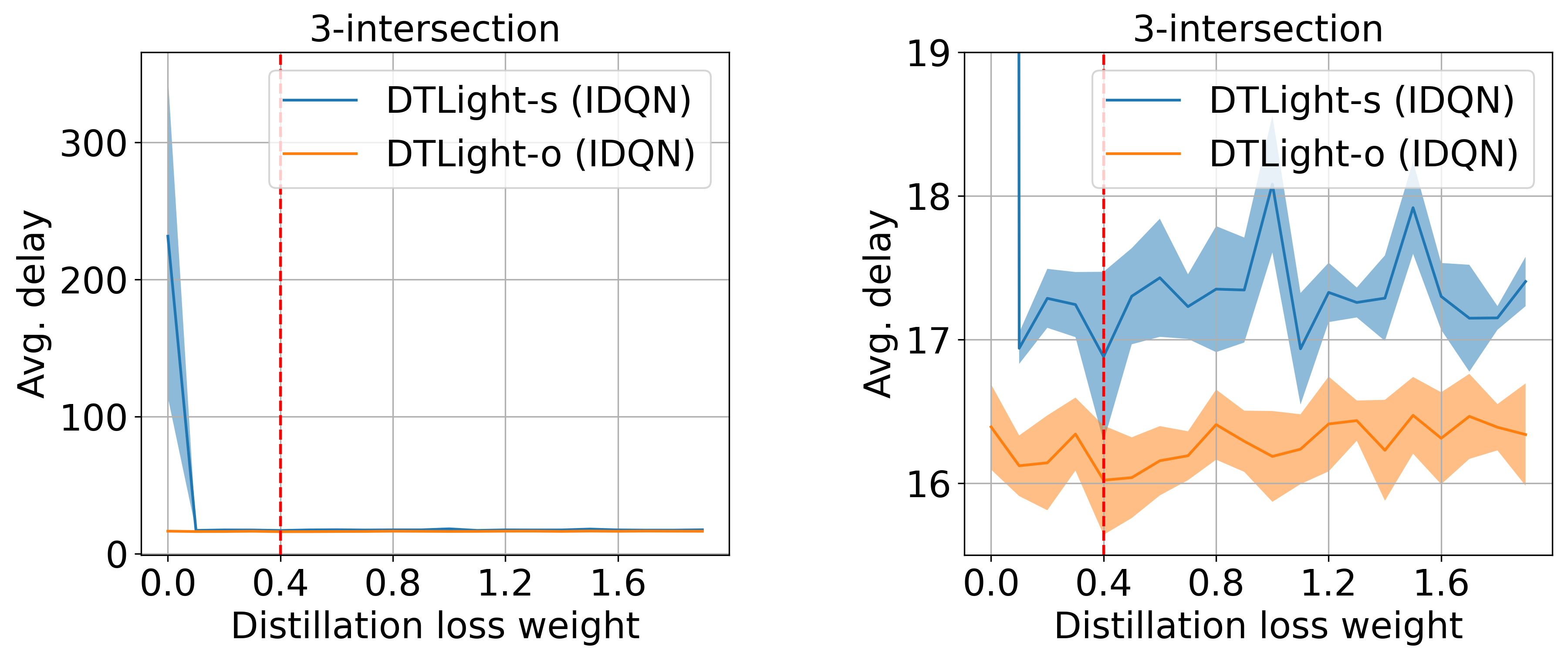}
\end{center}
\caption{
Performance of DTLight with different distillation loss weights $\alpha$ on the 3-intersection scenario. The right subfigure is a magnified view of the left one. 
}
\label{fig3}
\end{figure}

\begin{figure}
\begin{center}
\includegraphics[width=0.47\textwidth]{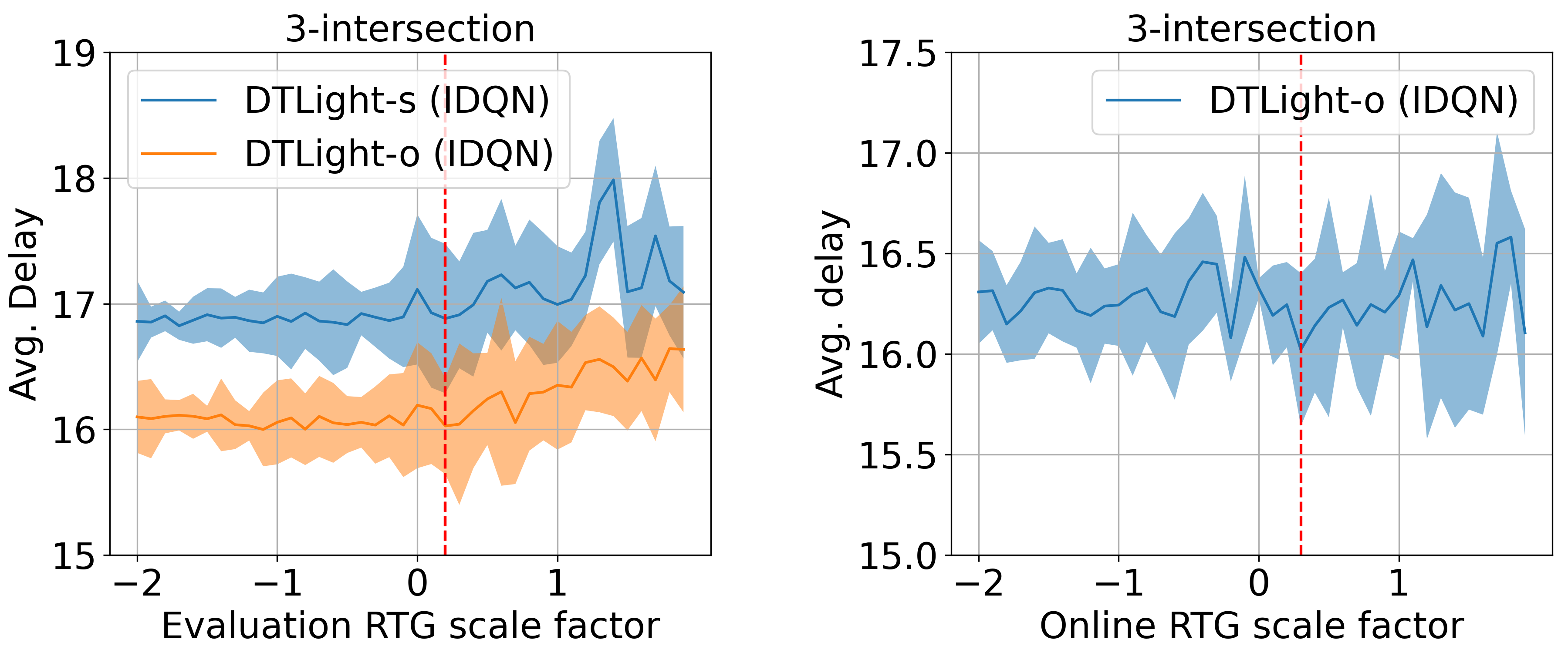}
\end{center}
\caption{Performance of DTLight with different evaluation and online RTG scale factors. The vertical red lines specify the chosen best RTG scale factor values.}
\label{fig4}
\end{figure}

\textbf{Distillation loss weight.}
Knowledge distillation significantly reduces the model size without compromising performance. This works by adopting an additional distillation loss during optimization. We compare the performance of two DTLight models with the same structure while pre-trained with different weights of only distillation loss given that we set the weight of hard loss $\beta$ to 1. As can be seen in Figure~\ref{fig3}, when the distill loss weight is zero, DTLight-s demonstrates an inability to learn an effective policy without the guidance of a teacher model. Nonetheless, through the implementation of online fine-tuning, DTLight's ability to attain a qualified policy is reinstated. We set the distillation loss weight $\alpha$ to 0.4 given that it achieves the lowest average delay for both pre-trained and finetuned models.

\textbf{RTG scale factor.}
RTG scale factors better exploit the offline datasets, freeing us from manually setting different RTG values for each DTLight agent. We demonstrate their impacts by varying the values $\gamma_{eval}$ and $\gamma_{online}$ for evaluation and online fine-tuning, respectively. The left panel in Figure~\ref{fig4} suggests that, for values exceeding 0.2, a reduction in $\gamma_{eval}$ leads to a lower average delay. Note that the return is negative in our TSC settings, thereby smaller scale factor indicates a higher initial RTG value. When the value is below 0.2, decreasing $\gamma_{eval}$ is unable to yield additional gains. As shown in the right panel of Figure\ref{fig4}, values of online RTG scale factor $\gamma_{online}$ above or below 0.3 result in worse delay.

\begin{figure}
\begin{center}
\includegraphics[width=0.48\textwidth]{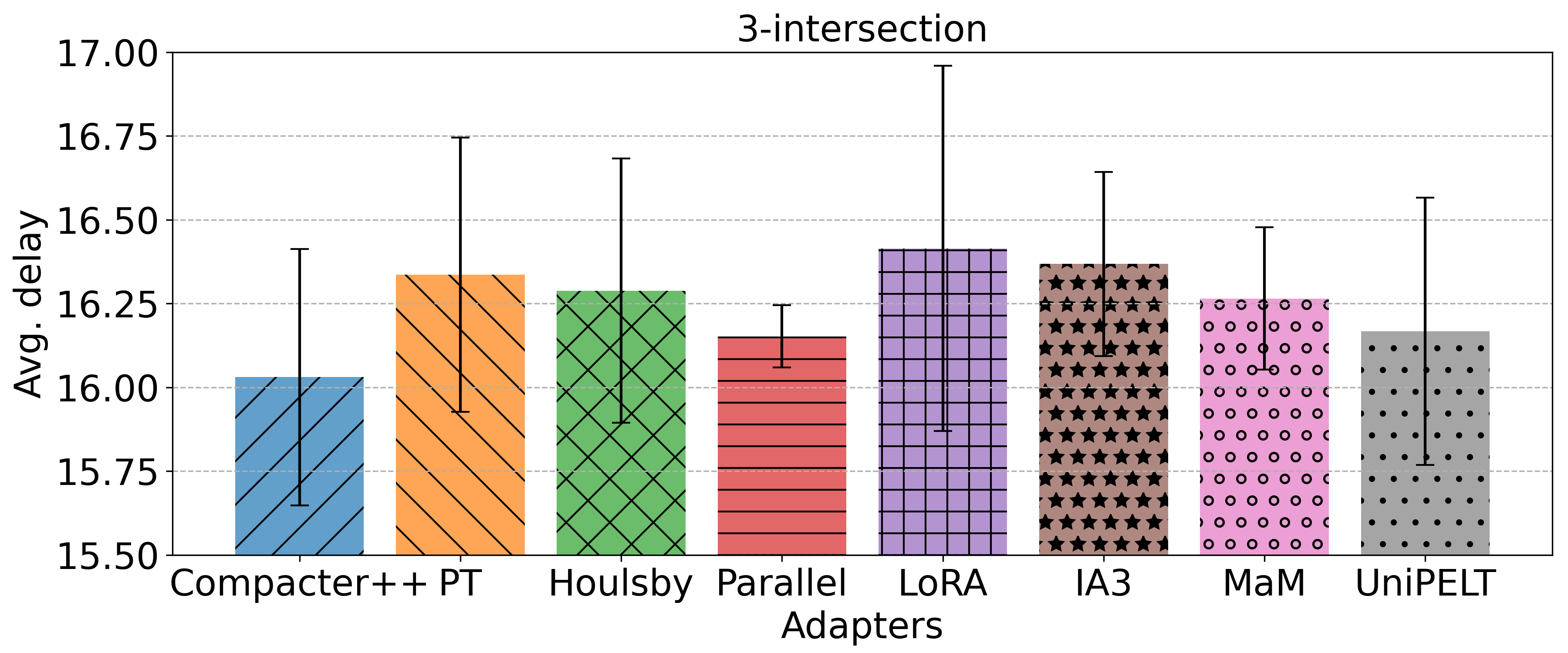}
\end{center}
\caption{Performance of DTLight with different adapter structures.}
\label{fig5}
\end{figure}

\textbf{Adapter module.}
We compare the COMPACTER++ adapter module we adopt~\citep{karimi2021compacter} with some other adapter structures from AdapterHub~\cite{pfeiffer2020AdapterHub}, i.e., Houlsby \citep{houlsby2019parameter}, Parallel \citep{he2021towards}, PrefixTuning (PT) \citep{li2021prefix}, LoRA \citep{hu2021lora}, IA3~\citep{liu2022few}, Mix-and-Match (MaM)~\citep{he2021towards}, and UniPELT~\citep{mao2021unipelt}. Incorporating COMPACTER++ as the adapter module in DTLight is motivated not only by its notable advantage in the number of learning parameters but also by its comparatively superior performance over other adapters, as demonstrated in Figure~\ref{fig5}.

\section{Conclusion and Discussion}
\label{Conclusion and Discussion}
This paper proposes DTLight, an offline-to-online method for both single- and multi-intersection traffic signal control. We build our DTLight model based on Decision Transfomer and utilize knowledge distillation to obtain lightweight pre-trained models. Adapter modules are further employed to make DTLight feasible for on-device online fine-tuning. Additionally, DTLight is extended to multi-agent control by incorporating neighborhood information in the state setting. We also create DTRL, the first offline dataset for traffic signal control using both synthetic and real-world scenarios with single and multiple signalized intersections. Extensive experiments demonstrate that DTLight pre-trained on DTRL outperforms baselines on the majority of control scenarios. The performance can be further enhanced with only 10 episodes of online fine-tuning, yielding an improvement of up to 42.6\%. In our future work, we plan to explore better coordination between multiple DTLight agents in offline settings.


\bibstyle{aaai24}

\end{document}